\documentclass[11pt]{article}

\usepackage{acl}
\usepackage{times}
\usepackage{latexsym}
\usepackage[T1]{fontenc}
\usepackage[utf8]{inputenc}
\usepackage{microtype}
\usepackage{graphicx}
\usepackage{booktabs}

\title{Rank, Don't Regress: A 279M Text Ranker Beats 2B and 7B VLMs on Referential Urdu GUI Grounding}

\author{Hamza Raziq Khan \quad Aimal Khan \\
  FAST NUCES, Peshawar, Pakistan \\
  \texttt{hamza.raziq@nu.edu.pk} \quad \texttt{aimalkhan9247@gmail.com}}

\begin{document}
\maketitle

\begin{abstract}
UI grounding asks a model to find the screen element that a user instruction refers to. Most current systems use a vision-language model (VLM) that reads a screenshot and outputs pixel coordinates. We study this task on Urdu mobile web pages, a setting with no published UI grounding results. Instead of regressing coordinates, we take the candidate elements from the page DOM and train a 279M-parameter cross-encoder (mDeBERTa-v3-base) to pick one. The click point is the centre of the picked element's box, so it is exact whenever the right element is picked. We built a frozen test set of 304 instructions from six Urdu news sites; 149 of them (49\%) are referential and never quote the full target text. Test domains were banned from training data before collection started. The ranker scores 73.4\% point-in-box accuracy, against 29.6\% for UI-TARS-2B and 23.7\% for Qwen2.5-VL-7B, both zero-shot. On the referential slice the ranker scores 65.8\%, against 30.2\% for a string-matching baseline that needs no training. In a controlled comparison at small scale, with training data drawn from the same pages, a coordinate-regression model of similar size scores 1.6\%. Adding element geometry as input tokens gives the best variant, at 75.7\%. The ranker's errors concentrate in two places: elements that share identical text, and ordinal instructions such as ``the fourth link''.
\end{abstract}

\section{Introduction}

A computer-use agent acts through a screen. Before it can click, it has to ground: given an instruction such as ``open the sports section'', it must find the element the instruction means. The current default is a vision-language model that reads a screenshot and generates a point~\cite{cheng2024seeclick,qin2025uitars,bai2025qwen25vl}. Grounding accuracy is the ceiling on everything the agent does next, and it is usually the weakest part of the pipeline~\cite{zheng2024seeact}.

Two facts pushed us away from that default. First, coordinate regression is fragile under layout change. In pilot runs for this project, Qwen2.5-VL-7B scored 34.1\% zero-shot on an 82-item pilot set of mobile pages; the same model fine-tuned on desktop screenshots fell to 11.0\%, because training and test pages laid out the same content at different widths. The readout itself is part of the problem: on COCO photos, Qwen3-VL-4B picks the correct cell 68.5\% of the time when the cells are named options, and 20.0\% when it must answer with pixel coordinates~\cite{frontera2026readout}. Second, the language we care about has no grounding data at all. Public grounding benchmarks cover English and Chinese interfaces~\cite{cheng2024seeclick,li2025screenspotpro}; for Urdu, a right-to-left language with tens of millions of speakers, we found no published UI grounding result, and hand-labelling click points for a new language is slow work.

A web page removes the labelling problem. Its DOM lists every clickable element with its text, attributes, and bounding box. The supervision is already there. The open question is the output space: should a model regress a point from pixels, or select one element from the page's own candidate list?

Selection is not a new idea. MindAct~\cite{deng2023mind2web} trains a small DeBERTa ranker over DOM candidates, but uses it as a pre-filter: the ranker shortlists elements and a large language model makes the final pick. Deng et al.\ report that the ranker on its own is not competitive on Mind2Web. Their pages are English desktop pages with about 1,135 candidate elements per page on average. Our pages are Urdu mobile pages with a median of 25 candidates. We test whether the conclusion survives the change of regime: a trained ranker, alone, as the complete grounder.

Our contributions are three:

\begin{enumerate}
\item \textbf{First UI grounding results for Urdu.} We report the first UI grounding (element localization) results for Urdu, measured on real mobile websites: a frozen test set of 304 instructions over six Urdu news sites, collected and locked before any training, with test domains excluded from training at the registrable-domain level. Half of the items are referential: the instruction describes the target without quoting its full text.
\item \textbf{The ranker alone is enough on mobile web.} A 279M cross-encoder used as the complete grounder, with no downstream model, scores 73.4\% against 29.6\% for UI-TARS-2B and 23.7\% for Qwen2.5-VL-7B. This reverses the ranker-alone finding of Mind2Web in the mobile regime.
\item \textbf{A controlled rank-versus-regress study at small scale.} Training data for both models comes from the same pages: DOM ranking scores 73.4\% and pixel coordinate regression scores 1.6\%. The difference is the output space, not the data source.~\cite{wu2026relavis} run the VLM-scale version of this comparison in pixel space; Section 2 separates the two.
\end{enumerate}

Section 2 places this work against prior grounding systems. Section 3 defines the task and the frozen protocol. Sections 4 and 5 describe the ranker and the comparison systems. Section 6 gives results, Section 7 takes the failures apart, and Sections 8 and 9 state limits and ethics.

\section{Related Work}

\paragraph{Visual grounding with VLMs.} SeeClick~\cite{cheng2024seeclick} showed that a VLM fine-tuned for grounding can drive GUI agents, and released the ScreenSpot benchmark. CogAgent~\cite{hong2024cogagent} added a high-resolution image encoder. UGround~\cite{gou2025uground} trained a grounder on large-scale web data. UI-TARS~\cite{qin2025uitars} trains a native agent model end to end, and Qwen2.5-VL~\cite{bai2025qwen25vl} is a general VLM with strong built-in grounding. All of these systems output coordinates. Their benchmarks are desktop, English or Chinese, and screenshot-based: ScreenSpot-Pro~\cite{li2025screenspotpro} targets high-resolution professional software, while OSWorld~\cite{xie2024osworld}, AndroidWorld~\cite{rawles2024androidworld}, WebArena~\cite{zhou2024webarena}, and VisualWebArena~\cite{koh2024visualwebarena} evaluate full tasks. Our zero-shot VLM baselines are UI-TARS-2B and Qwen2.5-VL-7B, the strongest open coordinate grounders available to us when the build ran, plus one recent mobile-focused grounder, MAI-UI-2B~\cite{zhou2026maiui} (Section 5). A second recent mobile grounder, Ferret-UI Lite~\cite{yang2025ferretuilite}, has no released code or weights and could not be evaluated (Section 5).

\paragraph{Coordinate-free and structured grounding.} GUI-Actor~\cite{wu2025guiactor} removes coordinate tokens and instead attends over image patches with a learned action head. It is coordinate-free, but it stays in pixel space and at VLM scale (7B). Mind2Web~\cite{deng2023mind2web} works from the DOM: a fine-tuned DeBERTa ranker (MindAct) keeps the top candidates and an LLM picks among them. The final decision in MindAct belongs to the LLM, and the paper reports that the small ranker alone performs far below the full pipeline. Prune4Web~\cite{zhang2025prune4web} has an LLM write small programs that prune the DOM before a downstream model acts; again, the structured step filters and something else decides. WinDOM~\cite{lichen2026windom} reads its training labels off the DOM for free, as we do, but spends them on a 2B regression grounder for a desktop web clone. Two 2026 studies set the terms for this paper.~\cite{chen2026lexicalcoupling} recast grounded steps from AndroidControl, MoTIF, and Mind2Web as same-screen ranking and showed that a fine-tuned ranker stays tied to visible labels: masking the target element's text drops its R@1 from 0.694 to 0.151. Their rankers are diagnostics run on existing English benchmarks; none is deployed as the grounder.~\cite{wu2026relavis} compare output spaces under control: at 7B VLM scale, on an identical 100K training set, a coordinate-free attention model beats a coordinate-based one by about 3.7 points across English desktop and web benchmarks. That comparison runs in pixel space at VLM scale. Ours runs in the DOM at 279M parameters, on data drawn from the same pages, and the gap is 71.8 points. Cross-encoder ranking itself is standard in text retrieval since monoBERT~\cite{nogueira2019monobert}: score each candidate against the query with full attention, take the best. We take the ranking mechanism from MindAct, credit it in full, and change its role and regime: mobile web, Urdu, roughly 25 candidates per page, and the ranker's top-1 is the answer.

\paragraph{Low-resource interfaces.} Multilingual agent evaluation has reached low-resource languages, at coarser grains than element localization. LoWAgent~\cite{monjur2026lowagent} evaluates web agents in English, Hindi, Bengali, Urdu, and Amharic on synthetic localized websites, and reports multi-step task success. VLURes~\cite{atuhurra2026vlures} benchmarks grounding in Urdu, but the task there is matching long texts to natural images, not interface elements. macOSWorld~\cite{yang2025macosworld} ships genuinely mirrored Arabic right-to-left interfaces, on desktop, scored by task success. A Persian vision-language-action dataset~\cite{pourasghari2026persianvla} covers instruction-to-plan grounding over photographs, an embodied setting. None of these reports UI grounding accuracy for Urdu, and none runs on real mobile websites. The blocker for UI grounding in a new language is annotation cost: each item needs a person to read an instruction and mark a box. For web pages, the DOM gives the boxes for free, which is what makes this study possible without an annotation budget.

\section{Task and Data}

\subsection{Task definition}

One item consists of a mobile page view (a 390$\times$844 screenshot and its DOM candidate list) and one Urdu instruction. A system outputs one point. The item is a hit if the point falls inside the gold element's box (point-in-box accuracy). For the ranker, the output point is the centre of the top-ranked candidate's box, so a hit is the same event as ranking the gold element first. For the VLMs and the coordinate control, the output point is whatever the model generates.

\subsection{Candidate extraction}

Candidates are the visible clickable elements of the view: links, buttons, inputs, and elements with button or link roles, filtered by the same visibility rules across all items. Each candidate record holds the element text, tag, aria-label, placeholder, title, link target, an id/class snippet, the heading of the nearest ancestor section, and the bounding box. The frozen views carry a median of 25 candidates.

\subsection{The frozen protocol}

Test domains were fixed and banned from training before any test page was collected. The ban is at registrable-domain level: if a domain is a test domain, none of its pages, subdomains, or sibling properties may appear in training. Domains that were tried as test sites but yielded zero usable items stayed banned.

Each view was captured once: screenshot and DOM in the same page load, stored together, write-once. News homepages change within hours, and a gold box collected on a later visit would point at a different page. Paired capture is what keeps the gold labels valid. The frozen set holds 160 views from six sites: Siasat (75 items), Dawn (75), Express (75), Daily Khabrain (67), Daily Mashriq (11), and Ummat (1). No site contributes more than 25\% of the items.

\subsection{Instruction generators}

Instructions come from five rule-based generators, listed in Table~\ref{tab:generators}. The explicit generator quotes the element's full text (``click \guillemotleft X\guillemotright''). The other four are referential: they describe the target without quoting its full text. Ordinal items name a position among same-tag candidates in DOM order (``the fourth link''). Partial items quote only the first three words of a long headline. Section items name the heading the element sits under. Function items refer to a control by its job (search box, menu button, login link, home link), found by deterministic attribute rules.

Every referential item passes an automatic uniqueness check: exactly one candidate in the view may satisfy the description under the generator's own rule, or the item is discarded. Each item records which generator produced it. The referential slice makes up 149 of 304 items (49.0\%). In 64 items, some other candidate in the view carries the same text as the target; those items are the source of the duplicate failures in Section 7.

\begin{table}[t]
\centering
\footnotesize
\setlength{\tabcolsep}{4pt}
\begin{tabular}{llrr}
\toprule
Generator & Type & Items & Share \\
\midrule
Explicit & quotes text & 155 & 51.0\% \\
Ordinal & referential & 101 & 33.2\% \\
Partial & referential & 31 & 10.2\% \\
Section & referential & 13 & 4.3\% \\
Function & referential & 4 & 1.3\% \\
\midrule
Total & & 304 & 100\% \\
\bottomrule
\end{tabular}
\caption{Frozen test set by instruction generator. Referential items never quote the full target text; partial items quote a three-word prefix.}
\label{tab:generators}
\end{table}

\subsection{Training pool}

Training pages come from 12 Urdu sites outside the ban list: 532 views and 4,216 positive items (2,774 explicit, 1,442 referential), produced by the same generators and the same uniqueness checks. Explicit items are capped at 16 per view so that easy quoted-text items do not swamp the pool. The pool is skewed: ARY News supplies 2,179 positives (51.7\%) and Urdu Wikipedia 1,049 (24.9\%), with BBC Urdu (381), Jang (282), and Geo Urdu (224) behind them. Validation holds out whole sites, never random rows; the validation split has 419 positives. A separate coordinate file holds the 1,539 explicit positives for which a screenshot was stored, with their gold centres, drawn from the same pages, for the control model in Section 5.

\subsection{Independent verification}

After the build, a separate script recomputed every number in this paper from the raw files rather than from stored summaries. It recounted all items, rechecked oracle presence for all 304 items (304/304 gold boxes match their frozen candidates exactly), confirmed zero registrable-domain overlap between training and test data and zero shared screenshot paths, and recounted every accuracy cell in Table~\ref{tab:main} from the per-item outputs. Eight items, spread across sites and generators, were also checked by eye against their screenshots. Every check passed.

\section{Method}

\subsection{The ranker}

The ranker is a cross-encoder. For one item, it scores every candidate in the view against the instruction and returns the argmax. Each score comes from one forward pass over the pair (instruction, candidate string), encoded jointly by mDeBERTa-v3-base~\cite{he2023debertav3} with a single-logit classification head. The model has 278.8M parameters. It reads no pixels: the screenshot exists for the VLM baselines and for human checking, and the ranker never sees it.

\subsection{Candidate strings}

A candidate is rendered as its text followed by bracketed fields. Table~\ref{tab:reprs} defines the four variants we trained. The \textsc{mindact\_text} variant is the MindAct-style representation: text and tag only. The \textsc{full} variant adds the attribute fields when they carry information the text does not, plus the parent section heading. \textsc{full\_noparent} removes the section heading. \textsc{full\_geo} adds the box as four numbers normalised by the viewport size.

\begin{table}[t]
\centering
\footnotesize
\setlength{\tabcolsep}{4pt}
\begin{tabular}{lll}
\toprule
Variant & Candidate string & Max len \\
\midrule
\textsc{mindact\_text} & text \texttt{[tag]} & 160 \\
\textsc{full} & text \texttt{[tag]} + \texttt{[ARIA]} & 224 \\
 & \texttt{[PH]} \texttt{[TITLE]} \texttt{[SEC]} & \\
\textsc{full\_noparent} & \textsc{full} without \texttt{[SEC]} & 224 \\
\textsc{full\_geo} & \textsc{full} + \texttt{[BOX x y w h]} & 224 \\
\bottomrule
\end{tabular}
\caption{Candidate representations. Attribute fields are included only when present and different from the element text. \texttt{[SEC]} is the parent section heading; \texttt{[BOX]} holds the normalised box coordinates.}
\label{tab:reprs}
\end{table}

\subsection{Training}

For each positive item, the gold candidate gets label 1 and up to eight other candidates from the same view, sampled at random, get label 0. One exception: for explicit items, a candidate whose normalised text equals the gold text is never used as a negative. It is a defensible answer to the quoted text, and punishing the model for it would teach the wrong lesson; Section 7 shows the cost of this ambiguity at test time.

We train with AdamW~\cite{loshchilov2019adamw} at learning rate 1e-5, weight decay 0.01, a linear schedule with 10\% warmup, two epochs, and seed 13. The effective batch is 32 (batch 32 at max length 160; batch 16 with two accumulation steps at max length 224). One practical detail cost us two early runs: the mDeBERTa-v3 checkpoint loads in fp16 by default in our stack, and fp16 training drove the loss to NaN. All reported models are loaded and trained in fp32, with steps skipped when the loss is non-finite.

\section{Experiments}

\subsection{Systems compared}

\paragraph{String-match heuristic.} No training. The rule extracts the quoted text from the instruction, picks a candidate whose normalised text matches it exactly (ties broken by larger area), and otherwise falls back to the candidate with the highest token overlap with the quote (ties by overlap ratio, then area). For referential instructions, which carry no quote, the rule has almost nothing to work with. This baseline prices the value of the quoted text alone.

\paragraph{Ranker variants.} The four variants of Section 4, all trained on the same training split of the pool in Section 3.5 (validation sites held out) with the same recipe.

\paragraph{UI-TARS-2B.} ByteDance's GUI-specialised VLM~\cite{qin2025uitars}, zero-shot. It receives the screenshot and the instruction in its standard prompt format and outputs a point in 0-to-1000 coordinates. All 304 outputs parsed.

\paragraph{Qwen2.5-VL-7B.} Alibaba's general VLM~\cite{bai2025qwen25vl}, zero-shot, outputting pixel coordinates for the 390$\times$844 view. 253 of 304 outputs parsed; the other 51 produced no usable point and count as misses.

\paragraph{MAI-UI-2B.} The 2B grounder from Alibaba's MAI-UI family~\cite{zhou2026maiui}, a GUI foundation model line released in December 2025 and built for real-world mobile and desktop interfaces. Zero-shot. We ran it in its native grounding mode: the documented MAI-UI grounding prompt, with the final point read from the model's answer block and its coordinates converted from the model's documented 0-to-999 grid, exactly as the official agent code normalises them. 220 of 304 outputs contained a usable point; the other 84 produced no coordinates at all and count as misses.

\paragraph{Ferret-UI Lite.} Apple's 3B on-device GUI agent~\cite{yang2025ferretuilite} could not be evaluated: Apple has released no code and no weights for it, and no official model repository exists, so there was nothing openly available to run zero-shot.

\paragraph{Coordinate control.} SmolVLM-500M-Instruct~\cite{marafioti2025smolvlm} fine-tuned with LoRA~\cite{hu2022lora} (rank 16, learning rate 2e-4, two epochs, effective batch 32) on the 1,539 coordinate rows from the same training pages. It reads the screenshot and generates the gold centre as text. This is the rank-versus-regress control: same pages, similar size, pixel input, coordinate output. All 304 outputs parsed.

\subsection{Metric and compute}

Every system is scored by point-in-box accuracy on the same 304 frozen items. We report overall accuracy, the explicit slice (155 items), the referential slice (149 items), and per-generator accuracy. Ranker training, the ablations, both VLM evaluations, and the control training ran as separate jobs on a single Tesla T4: about 1.1 hours for the headline ranker, 2.0 hours for the two ablation variants, 52 minutes for both VLM evaluations, and 4.2 hours for the coordinate control. The MAI-UI-2B evaluation ran later, as a separate job on the same GPU.

\section{Results}

Table~\ref{tab:main} gives the main result. The full ranker scores 73.4\% overall. UI-TARS-2B scores 29.6\% and Qwen2.5-VL-7B 23.7\%. The gap to the stronger VLM is 43.8 points. MAI-UI-2B, the newest VLM we tested, scores 11.8\%, below both older VLMs; Section 5 gives its protocol and parse rate.

\begin{table}[t]
\centering
\footnotesize
\setlength{\tabcolsep}{3pt}
\begin{tabular}{lrrr}
\toprule
System & Overall & Explicit & Referential \\
\midrule
Heuristic (no training) & 57.9 & 84.5 & 30.2 \\
\textsc{mindact\_text} ranker & 74.0 & 80.6 & 67.1 \\
\textbf{\textsc{full} ranker} & \textbf{73.4} & \textbf{80.6} & \textbf{65.8} \\
\textsc{full\_noparent} & 72.7 & 80.6 & 64.4 \\
\textsc{full\_geo} & \textbf{75.7} & \textbf{86.5} & 64.4 \\
UI-TARS-2B & 29.6 & 40.0 & 18.8 \\
Qwen2.5-VL-7B & 23.7 & 32.9 & 14.1 \\
MAI-UI-2B & 11.8 & 13.5 & 10.1 \\
SmolVLM-500M control & 1.6 & 0.0 & 3.4 \\
\bottomrule
\end{tabular}
\caption{Point-in-box accuracy (\%) on the frozen test set: 304 items, 155 explicit and 149 referential. All VLM rows are zero-shot. Hit counts: heuristic 176, \textsc{mindact\_text} 225, \textsc{full} 223, \textsc{full\_noparent} 221, \textsc{full\_geo} 230, UI-TARS-2B 90, Qwen2.5-VL-7B 72, MAI-UI-2B 36, coordinate control 5. MAI-UI-2B parsed 220 of its 304 outputs; the rest contained no point and count as misses. Ferret-UI Lite has no row: it could not be evaluated, for the reason given in Section 5.}
\label{tab:main}
\end{table}

One property of the test set must be stated next to these numbers. Half of the items are referential by design, and the VLMs are weakest exactly there: on ordinal items they score 4.0\% and 2.0\% (Table~\ref{tab:pergen}). On the explicit slice, where the instruction quotes the text, they reach 40.0\% and 32.9\%. Their low overall scores partly reflect the slice mix, and we report the slices so the reader can see that split directly. The comparison that does not depend on the mix is the control: a coordinate model trained on the same pages scores 1.6\% overall and 0 of 155 on the explicit slice, where the answer is printed on the screen in the instruction itself.

For scale, Wilson 95\% confidence intervals on the headline numbers are [68.1, 78.0] for the full ranker overall, [24.8, 35.0] for UI-TARS-2B, and [19.3, 28.8] for Qwen2.5-VL-7B. On the referential slice the full ranker sits at [57.8, 72.9] against the heuristic's [23.4, 38.0]. None of the intervals behind the three main claims overlap.

\subsection{Where the win comes from}

The heuristic wins the explicit slice outright, 84.5\% against 80.6\% for the rankers. That is expected: the instruction quotes the target text, and exact matching is the right tool. The ranking claim rests on the referential slice. There, the full ranker scores 65.8\% against the heuristic's 30.2\%, a gap of 35.6 points on items that never quote the full answer. On ordinal items the heuristic scores 1.0\% and the ranker 49.5\%. A model that had only learned to match strings could not produce these numbers; the ranker learned to use tag, attributes, section context, and position cues that the heuristic cannot read.

\begin{table*}[t]
\centering
\small
\begin{tabular}{lrrrrr}
\toprule
System & Explicit (155) & Ordinal (101) & Partial (31) & Section (13) & Function (4) \\
\midrule
Heuristic & 84.5 & 1.0 & 100 & 100 & 0.0 \\
\textsc{mindact\_text} & 80.6 & 51.5 & 100 & 100 & 100 \\
\textsc{full} & 80.6 & 49.5 & 100 & 100 & 100 \\
\textsc{full\_noparent} & 80.6 & 50.5 & 100 & 100 & 25.0 \\
\textsc{full\_geo} & 86.5 & 47.5 & 100 & 100 & 100 \\
UI-TARS-2B & 40.0 & 4.0 & 48.4 & 46.2 & 75.0 \\
Qwen2.5-VL-7B & 32.9 & 2.0 & 48.4 & 7.7 & 75.0 \\
MAI-UI-2B & 13.5 & 1.0 & 41.9 & 7.7 & 0.0 \\
Coord.\ control & 0.0 & 3.0 & 6.5 & 0.0 & 0.0 \\
\bottomrule
\end{tabular}
\caption{Accuracy (\%) by instruction generator. The function slice has four items, so its percentages move in steps of 25. The ordinal slice is where the systems separate.}
\label{tab:pergen}
\end{table*}

\begin{figure*}[t]
\centering
\begin{minipage}[t]{0.48\textwidth}
\centering
\includegraphics[width=\linewidth]{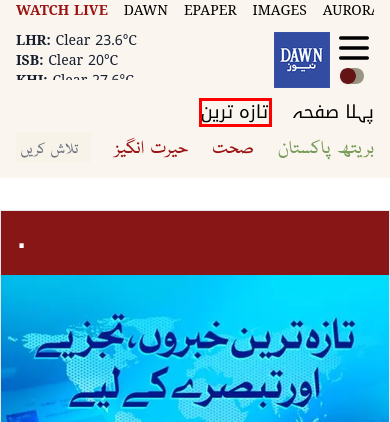}\\
(a) Explicit item, Dawn
\end{minipage}
\hfill
\begin{minipage}[t]{0.48\textwidth}
\centering
\includegraphics[width=\linewidth]{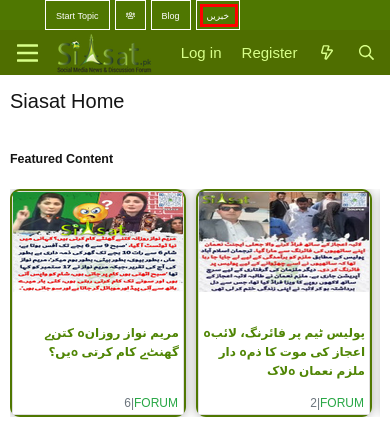}\\
(b) Ordinal item, Siasat
\end{minipage}
\caption{Two frozen test items: one the string-matching baseline can solve, one it cannot. The screenshots are cropped to the top half so the gold boxes stay legible; both targets sit in the page header, marked by the red boxes. (a) The instruction quotes the target text (``click `Latest''). (b) The instruction (``click the fourth link'') quotes no text, and the count runs over links in DOM order.}
\label{fig:examples}
\end{figure*}

Figure~\ref{fig:examples} shows the two ends of the difficulty range: one explicit item and one ordinal item.

\subsection{Ablations}

The richer candidate string did not help. \textsc{mindact\_text}, with text and tag only, scores 74.0\% overall and 67.1\% on the referential slice, slightly ahead of \textsc{full} at 73.4\% and 65.8\%. On 304 items a 0.6-point difference is two items, and we do not read anything into it beyond this: attributes and section headings added no measurable gain on this data. Removing the section heading from \textsc{full} (\textsc{full\_noparent}, 72.7\%) costs 0.7 points, consistent with that reading.

Geometry did help, in one place. \textsc{full\_geo} scores 75.7\% overall, the best result in the study, and the gain sits in the explicit slice: 86.5\% against 80.6\%, while the referential slice stays at 64.4\%. Explicit errors are mostly duplicate picks (Section 7), and a box position is exactly the signal that separates two elements with the same text.

Held-out-site validation during training tells a different ordering: ranking accuracy on held-out training sites is 68.5\% for \textsc{full}, 61.1\% for \textsc{full\_geo}, 60.1\% for \textsc{mindact\_text}, and 58.7\% for \textsc{full\_noparent} (419 validation positives). The validation ranking and the frozen-test ranking disagree, so we report all four variants and base our claims on the \textsc{full} configuration fixed before the ablation runs, not on the test-set winner.

\subsection{Latency and cost}

Table~\ref{tab:latency} lists measured per-item times, with the hardware stated for each row, because the rows were not measured on the same machine. The rankers ran in fp32 on a shared 2-core VM CPU; the VLMs and the control ran on a Tesla T4 GPU. On this hardware the full ranker, at a median of 7.44 seconds per query over 31 candidates, is slower per item than UI-TARS-2B served on a GPU (2.61 seconds). We make no wall-clock claim. What the ranker offers is 279M parameters, text-only input, training in about an hour on one GPU, and inference that runs on a CPU at all. A phone-class agent that cannot host a 2B or 7B VLM can host this ranker; whether it is fast enough there is a separate measurement we have not done.

\begin{table}[!tb]
\centering
\small
\begin{tabular}{lllr}
\toprule
System & Params & Device & s/item \\
\midrule
\textsc{mindact\_text} & 279M & CPU, 2-core VM & 6.07 \\
\textsc{full} & 279M & CPU, 2-core VM & 7.44 \\
UI-TARS-2B & 2B & GPU, Tesla T4 & 2.61 \\
Qwen2.5-VL-7B & 7B & GPU, Tesla T4 & 5.68 \\
MAI-UI-2B & 2B & GPU, Tesla T4 & 13.20 \\
Coord.\ control & 500M & GPU, Tesla T4 & 2.39 \\
\bottomrule
\end{tabular}
\caption{Median seconds per item. Ranker times are medians over 49 frozen items with a median of 31 candidates per view, fp32, no batching or quantisation. VLM and control times are per screenshot on GPU. The devices differ, so read this table as a cost profile, not a race.}
\label{tab:latency}
\end{table}

\section{Analysis}

We classified all 81 failures of the full ranker from its per-item outputs. The failures fall into exactly two groups.

\paragraph{Same-text duplicates (30 failures).} Every explicit failure picked a different element carrying the same text as the target. News navigation repeats labels across a page (a section name appears in the top bar, the side rail, and the footer), and 64 of the 304 items have a duplicated target text. When the duplicates also share a section context, the candidate strings are near-identical and the pick is close to a coin flip. The geometry variant cuts these errors, as Section 6 shows, because position is one signal the duplicates do not share. One catalogued case (item it0275, Express): the instruction quotes the label ``Business'', and the element the ranker picked carries the identical label. Only the position is wrong.

\paragraph{Ordinal confusion (51 failures).} Every referential failure is an ordinal item: the ranker got 50 of 101 ordinal items right and spread the other 51 over wrong positions. Partial, section, and function items are error-free (48 of 48). The cause is structural. Each candidate is scored in a separate forward pass, so the model never sees the list as a list; nothing in a candidate string says ``I am the fourth link''. Position has to be inferred from the box coordinates and the surrounding fields, and \textsc{full\_geo} does not fix it either (47.5\% on ordinal items). In item it0290 (Express), the instruction asks for the third link; the target is the ``Live TV'' link, and the ranker picked a different link, labelled ``Aap Ka Din''. Ordinal grounding needs the candidates scored jointly, or an explicit position feature that means something. We did neither in this build.

\paragraph{The lexical-coupling test.}~\cite{chen2026lexicalcoupling} showed that a fine-tuned ranker on English benchmarks is, to a large degree, recovering visible labels: mask the target element's text and R@1 falls from 0.694 to 0.151. If our ranker worked that way, it should collapse on instructions that never contain the label. The referential slice is built to be exactly that test, and the ordinal slice is its hardest part, since the instruction carries a number and a tag word and nothing else. Pure label recovery is in the table: the string-match heuristic scores 30.2\% on the referential slice and 1.0\% on ordinal items. The full ranker scores 65.8\% and 49.5\%. Between the full set and the referential slice, the ranker's accuracy drops 7.6 points (73.4 to 65.8); the heuristic's drops 27.7 (57.9 to 30.2). We do not claim the ranker is free of lexical coupling. It still loses the explicit slice to exact string matching (80.6\% against 84.5\%), and its duplicate failures are lexical at root. The narrower claim is what the slices support: when the instruction withholds the label, learned ranking keeps most of its accuracy and label matching keeps almost none of its.

\paragraph{One hard site.} Express contributes 75 items and the full ranker gets 26 of them (34.7\%), against 82.7\% on Dawn, 85.3\% on Siasat, and 95.5\% on Daily Khabrain. Table~\ref{tab:persite} gives accuracy by site for every system. The site's item mix stacks both failure modes: 40 of its 75 items are ordinal and 31 have a duplicated target text. The per-site gap is a property of the items, not a site-specific bug we found: the VLMs collapse on the same site (UI-TARS-2B 4.0\%, Qwen2.5-VL-7B 0.0\%).

\begin{table*}[!t]
\centering
\footnotesize
\setlength{\tabcolsep}{4pt}
\begin{tabular}{lrrrrrr}
\toprule
System & Siasat (75) & Dawn (75) & Express (75) & Khabrain (67) & Mashriq (11) & Ummat (1) \\
\midrule
Heuristic & 42.7 & 76.0 & 32.0 & 83.6 & 63.6 & 0.0 \\
\textsc{mindact\_text} & 85.3 & 81.3 & 37.3 & 95.5 & 63.6 & 100 \\
\textsc{full} & 85.3 & 82.7 & 34.7 & 95.5 & 63.6 & 0.0 \\
\textsc{full\_noparent} & 85.3 & 78.7 & 36.0 & 95.5 & 63.6 & 0.0 \\
\textsc{full\_geo} & 90.7 & 82.7 & 36.0 & 95.5 & 81.8 & 0.0 \\
UI-TARS-2B & 29.3 & 49.3 & 4.0 & 34.3 & 45.5 & 0.0 \\
Qwen2.5-VL-7B & 25.3 & 42.7 & 0.0 & 28.4 & 18.2 & 0.0 \\
MAI-UI-2B & 13.3 & 16.0 & 2.7 & 14.9 & 18.2 & 0.0 \\
Coord.\ control & 0.0 & 4.0 & 1.3 & 1.5 & 0.0 & 0.0 \\
\bottomrule
\end{tabular}
\caption{Accuracy (\%) by test site, item counts in parentheses. Mashriq and Ummat are too small to read on their own; the four large sites carry the comparison. Express is the hardest site for every system, including both VLMs.}
\label{tab:persite}
\end{table*}

\paragraph{The control's failure mode.} The coordinate control parsed cleanly (304 of 304 outputs) and still scored 1.6\%. Its errors are spread points: with 1,539 training rows it did not learn where Urdu page elements sit, and its 0 of 155 on the explicit slice shows the problem is not instruction understanding. Given the screenshot and the exact text to find, it cannot point at the text. That is the regression task failing on its own terms, at the same data scale where ranking reaches 73.4\%.

\section{Limitations}

The test set covers six sites, all news. Forms, shops, banking, and government portals have different layouts and different element mixes, and we have no numbers for them.

The instructions are rule-generated. Five generators produce clean, unambiguous Urdu, and every referential item passed a uniqueness check by construction. Instructions written by people will be messier, and our accuracy on them is unmeasured.

The test set is small. At 304 items, one item is worth about 0.33 points, and differences between ranker variants of under two points should be read as noise. The claims we make rest on gaps of 35 points or more.

Two weaknesses are measured, not hypothetical. When two elements carry identical text and similar context, the ranker often picks the wrong copy (30 of its 81 failures). Ordinal instructions stay near 50\% for the ranker and under 5\% for both VLMs, because no system we tested represents list position well.

The method needs the DOM. It applies to web pages rendered in a browser or webview. It does not apply to native apps, canvas interfaces, or a bare screenshot, and in those settings coordinate models remain the only option of the two.

The training pool is skewed. Two sites supply 76.6\% of the positives, so the ranker may be fitted to ARY News and Wikipedia layouts more than to Urdu pages in general. The frozen test sites are disjoint from training, which limits how much this can inflate the test numbers, but it can still narrow what the model learned.

Latency was measured on a shared 2-core VM in fp32, with no batching, quantisation, or distillation. Those numbers describe our measurement setup, not a deployment.

\section{Ethics}

All pages came from public news sites, fetched without accounts or logins. The records hold page structure, element text, and boxes; they contain no personal data, no user accounts, and no tracking of any person. Ad and tracker requests were blocked during collection. The instructions are machine-generated; no human subjects took part and no annotation labour was used.

Grounding models are components of web automation, and automation can be misused, for example to operate accounts or submit content at scale. Our study is read-only: the systems predict points on saved pages, and no action was performed on any live site during data collection or evaluation.

Page text and layout belong to the publishers. We use short element strings and bounding boxes for research evaluation, and the figures show single elements in context for the same purpose.

\section{Conclusion}

On a frozen Urdu mobile-web test set, a 279M text ranker used on its own scores 73.4\%, against 29.6\% and 23.7\% for 2B and 7B VLMs. MindAct's ranking mechanism is not ours; what the numbers show is that its role can change when the page carries about 25 candidates instead of about a thousand, and when the interface language is Urdu, where both VLMs score under 30\%. The ranker then needs no LLM behind it and no GPU in front of a user. The controlled comparison makes the same point from the other side: at small scale, with both models trained on data from the same pages, regressing coordinates scores 1.6\% where DOM ranking scores 73.4\%. What is left open is clear from the failures: score candidates jointly so list position exists, use local context to separate same-text duplicates, and test on human-written instructions, other genres, and a second right-to-left language.

\bibliography{references}

\begin{thebibliography}{29}
\providecommand{\natexlab}[1]{#1}

\bibitem[{Atuhurra et~al.(2026)Atuhurra, Ali, Iwakura, Kamigaito, and
  Hiraoka}]{atuhurra2026vlures}
Jesse Atuhurra, Iqra Ali, Tomoya Iwakura, Hidetaka Kamigaito, and Tatsuya
  Hiraoka. 2026.
\newblock {VLURes}: Benchmarking long-text grounding and cross-lingual
  robustness in vision language models.
\newblock In \emph{Findings of the Association for Computational Linguistics:
  ACL 2026}, pages 27426--27481.

\bibitem[{Bai et~al.(2025)Bai, Chen, Liu, Wang, Ge, Song, Dang, Wang, Wang,
  Tang, Zhong, Zhu, Yang, Li, Wan, Wang, Ding, Fu, Xu, Ye, Zhang, Xie, Cheng,
  Zhang, Yang, Xu, and Lin}]{bai2025qwen25vl}
Shuai Bai, Keqin Chen, Xuejing Liu, Jialin Wang, Wenbin Ge, Sibo Song, Kai
  Dang, Peng Wang, Shijie Wang, Jun Tang, Humen Zhong, Yuanzhi Zhu, Mingkun
  Yang, Zhaohai Li, Jianqiang Wan, Pengfei Wang, Wei Ding, Zheren Fu, Yiheng
  Xu, and 8 others. 2025.
\newblock {Qwen2.5-VL} technical report.
\newblock \emph{arXiv preprint arXiv:2502.13923}.

\bibitem[{Chen and Jacucci(2026)}]{chen2026lexicalcoupling}
Qijia Chen and Giulio Jacucci. 2026.
\newblock Lexical coupling in {GUI} element grounding: Sentence embeddings
  track labels across mobile and web.
\newblock \emph{arXiv preprint arXiv:2608.21794}.

\bibitem[{Cheng et~al.(2024)Cheng, Sun, Chu, Xu, Li, Zhang, and
  Wu}]{cheng2024seeclick}
Kanzhi Cheng, Qiushi Sun, Yougang Chu, Fangzhi Xu, Yantao Li, Jianbing Zhang,
  and Zhiyong Wu. 2024.
\newblock {SeeClick}: Harnessing {GUI} grounding for advanced visual {GUI}
  agents.
\newblock In \emph{Proceedings of the 62nd Annual Meeting of the Association
  for Computational Linguistics (ACL)}.

\bibitem[{Deng et~al.(2023)Deng, Gu, Zheng, Chen, Stevens, Wang, Sun, and
  Su}]{deng2023mind2web}
Xiang Deng, Yu~Gu, Boyuan Zheng, Shijie Chen, Samuel Stevens, Boshi Wang, Huan
  Sun, and Yu~Su. 2023.
\newblock {Mind2Web}: Towards a generalist agent for the web.
\newblock In \emph{Advances in Neural Information Processing Systems
  (NeurIPS)}.

\bibitem[{Frontera Del~Valle(2026)}]{frontera2026readout}
Alfredo~F. Frontera Del~Valle. 2026.
\newblock What looks like a capability limit in vision-language models is a
  readout limit.
\newblock \emph{arXiv preprint arXiv:2609.27408}.

\bibitem[{Gou et~al.(2025)Gou, Wang, Zheng, Xie, Chang, Shu, Sun, and
  Su}]{gou2025uground}
Boyu Gou, Ruohan Wang, Boyuan Zheng, Yanan Xie, Cheng Chang, Yiheng Shu, Huan
  Sun, and Yu~Su. 2025.
\newblock Navigating the digital world as humans do: Universal visual grounding
  for {GUI} agents.
\newblock In \emph{International Conference on Learning Representations
  (ICLR)}.

\bibitem[{He et~al.(2023)He, Gao, and Chen}]{he2023debertav3}
Pengcheng He, Jianfeng Gao, and Weizhu Chen. 2023.
\newblock {DeBERTaV3}: Improving {DeBERTa} using {ELECTRA}-style pre-training
  with gradient-disentangled embedding sharing.
\newblock In \emph{International Conference on Learning Representations
  (ICLR)}.

\bibitem[{Hong et~al.(2024)Hong, Wang, Lv, Xu, Yu, Ji, Wang, Wang, Dong, Ding,
  and Tang}]{hong2024cogagent}
Wenyi Hong, Weihan Wang, Qingsong Lv, Jiazheng Xu, Wenmeng Yu, Junhui Ji, Yan
  Wang, Zihan Wang, Yuxiao Dong, Ming Ding, and Jie Tang. 2024.
\newblock {CogAgent}: A visual language model for {GUI} agents.
\newblock In \emph{Proceedings of the IEEE/CVF Conference on Computer Vision
  and Pattern Recognition (CVPR)}, pages 14281--14290.

\bibitem[{Hu et~al.(2022)Hu, Shen, Wallis, Allen-Zhu, Li, Wang, Wang, and
  Chen}]{hu2022lora}
Edward~J. Hu, Yelong Shen, Phillip Wallis, Zeyuan Allen-Zhu, Yuanzhi Li, Shean
  Wang, Lu~Wang, and Weizhu Chen. 2022.
\newblock {LoRA}: Low-rank adaptation of large language models.
\newblock In \emph{International Conference on Learning Representations
  (ICLR)}.

\bibitem[{Koh et~al.(2024)Koh, Lo, Jang, Duvvur, Lim, Huang, Neubig, Zhou,
  Salakhutdinov, and Fried}]{koh2024visualwebarena}
Jing~Yu Koh, Robert Lo, Lawrence Jang, Vikram Duvvur, Ming~Chong Lim, Po-Yu
  Huang, Graham Neubig, Shuyan Zhou, Ruslan Salakhutdinov, and Daniel Fried.
  2024.
\newblock {VisualWebArena}: Evaluating multimodal agents on realistic visual
  web tasks.
\newblock In \emph{Proceedings of the 62nd Annual Meeting of the Association
  for Computational Linguistics (ACL)}.

\bibitem[{Li et~al.(2025)Li, Meng, Lin, Luo, Tian, Ma, Huang, and
  Chua}]{li2025screenspotpro}
Kaixin Li, Ziyang Meng, Hongzhan Lin, Ziyang Luo, Yuchen Tian, Jing Ma, Zhiyong
  Huang, and Tat-Seng Chua. 2025.
\newblock {ScreenSpot-Pro}: {GUI} grounding for professional high-resolution
  computer use.
\newblock \emph{arXiv preprint arXiv:2504.07981}.

\bibitem[{Li-Chen et~al.(2026)Li-Chen, Zhou, Chen, and
  Chauvin}]{lichen2026windom}
Chengheng Li-Chen, Zhiqian Zhou, Hao Chen, and Nicolas Chauvin. 2026.
\newblock {WinDOM}: Self-family distillation for small-model {GUI} grounding.
\newblock \emph{arXiv preprint arXiv:2606.25964}.

\bibitem[{Loshchilov and Hutter(2019)}]{loshchilov2019adamw}
Ilya Loshchilov and Frank Hutter. 2019.
\newblock Decoupled weight decay regularization.
\newblock In \emph{International Conference on Learning Representations
  (ICLR)}.

\bibitem[{Marafioti et~al.(2025)Marafioti, Zohar, Farr{\'e}, Noyan, Bakouch,
  Cuenca, Zakka, Ben~Allal, Lozhkov, Tazi, Srivastav, Lochner, Larcher, Morlon,
  Tunstall, von Werra, and Wolf}]{marafioti2025smolvlm}
Andr{\'e}s Marafioti, Orr Zohar, Miquel Farr{\'e}, Merve Noyan, Elie Bakouch,
  Pedro Cuenca, Cyril Zakka, Loubna Ben~Allal, Anton Lozhkov, Nouamane Tazi,
  Vaibhav Srivastav, Joshua Lochner, Hugo Larcher, Mathieu Morlon, Lewis
  Tunstall, Leandro von Werra, and Thomas Wolf. 2025.
\newblock {SmolVLM}: Redefining small and efficient multimodal models.
\newblock \emph{arXiv preprint arXiv:2504.05299}.

\bibitem[{Monjur et~al.(2026)Monjur, Nisar, Nahin, Belay, Mishra, Azime, and
  Chhabra}]{monjur2026lowagent}
Ocean Monjur, Daud~Ahmad Nisar, Shahriar~Kabir Nahin, Tadesse~Destaw Belay,
  Sushruti Mishra, Israel~Abebe Azime, and Anshuman Chhabra. 2026.
\newblock {LoWAgent}: A low-resource multilingual benchmark for web agents.
\newblock \emph{arXiv preprint arXiv:2610.27205}.

\bibitem[{Nogueira and Cho(2019)}]{nogueira2019monobert}
Rodrigo Nogueira and Kyunghyun Cho. 2019.
\newblock Passage re-ranking with {BERT}.
\newblock \emph{arXiv preprint arXiv:1901.04085}.

\bibitem[{Pourasghari et~al.(2026)Pourasghari, Sadidpour, and
  Roustaei}]{pourasghari2026persianvla}
Abdollah Pourasghari, Saeedeh~Sadat Sadidpour, and Mohammad Roustaei. 2026.
\newblock A {Persian} vision-language-action dataset and benchmark for
  ambiguity- and infeasibility-aware instruction-to-plan grounding.
\newblock \emph{arXiv preprint arXiv:2610.15666}.

\bibitem[{Qin et~al.(2025)Qin, Ye, Fang, Wang, Liang, Tian, Zhang, Li, Li,
  Huang, Zhong, Li et~al.}]{qin2025uitars}
Yujia Qin, Yining Ye, Junjie Fang, Haoming Wang, Shihao Liang, Shizuo Tian,
  Junda Zhang, Jiahao Li, Yunxin Li, Shijue Huang, Wanjun Zhong, Kuanye Li,
  et~al. 2025.
\newblock {UI-TARS}: Pioneering automated {GUI} interaction with native agents.
\newblock \emph{arXiv preprint arXiv:2501.12326}.

\bibitem[{Rawles et~al.(2024)Rawles, Clinckemaillie, Chang, Waltz, Lau, Fair,
  Li, Bishop, Li, Campbell-Ajala, Toyama, Berry, Tyamagundlu, Lillicrap, and
  Riva}]{rawles2024androidworld}
Christopher Rawles, Sarah Clinckemaillie, Yifan Chang, Jonathan Waltz,
  Gabrielle Lau, Marybeth Fair, Alice Li, William Bishop, Wei Li, Folawiyo
  Campbell-Ajala, Daniel Toyama, Robert~J. Berry, Divya Tyamagundlu, Timothy
  Lillicrap, and Oriana Riva. 2024.
\newblock {AndroidWorld}: A dynamic benchmarking environment for autonomous
  agents.
\newblock \emph{arXiv preprint arXiv:2405.14573}.

\bibitem[{Wu et~al.(2026)Wu, Zheng, Huang, and Wang}]{wu2026relavis}
Kaiyu Wu, Beichen Zheng, Weiyao Huang, and Keze Wang. 2026.
\newblock Recurrent latent visual search for {GUI} grounding.
\newblock \emph{arXiv preprint arXiv:2610.05185}.

\bibitem[{Wu et~al.(2025)Wu, Cheng, Yang, Zhang, Yang, Jiang, Mu, Peng, Qiao,
  Tan, Qin, Liden, Lin, Zhang, Zhang, Zhang, Zhang, and Gao}]{wu2025guiactor}
Qianhui Wu, Kanzhi Cheng, Rui Yang, Chaoyun Zhang, Jianwei Yang, Huiqiang
  Jiang, Jian Mu, Baolin Peng, Bo~Qiao, Reuben Tan, Si~Qin, Lars Liden, Qingwei
  Lin, Huan Zhang, Tong Zhang, Jianbing Zhang, Dongmei Zhang, and Jianfeng Gao.
  2025.
\newblock {GUI-Actor}: Coordinate-free visual grounding for {GUI} agents.
\newblock \emph{arXiv preprint arXiv:2506.03143}.

\bibitem[{Xie et~al.(2024)Xie, Zhang, Chen, Li, Zhao, Cao, Hua, Cheng, Shin,
  Lei, Liu, Xu, Zhou, Savarese, Xiong, Zhong, and Yu}]{xie2024osworld}
Tianbao Xie, Danyang Zhang, Jixuan Chen, Xiaochuan Li, Siheng Zhao, Ruisheng
  Cao, Toh~Jing Hua, Zhoujun Cheng, Dongchan Shin, Fangyu Lei, Yitao Liu,
  Yiheng Xu, Shuyan Zhou, Silvio Savarese, Caiming Xiong, Victor Zhong, and Tao
  Yu. 2024.
\newblock {OSWorld}: Benchmarking multimodal agents for open-ended tasks in
  real computer environments.
\newblock In \emph{Advances in Neural Information Processing Systems
  (NeurIPS)}.

\bibitem[{Yang et~al.(2025{\natexlab{a}})Yang, Ci, and
  Shou}]{yang2025macosworld}
Pei Yang, Hai Ci, and Mike~Zheng Shou. 2025{\natexlab{a}}.
\newblock {macOSWorld}: A multilingual interactive benchmark for {GUI} agents.
\newblock \emph{arXiv preprint arXiv:2506.04135}.

\bibitem[{Yang et~al.(2025{\natexlab{b}})Yang, Dou, Feng, Huang, Nguyen, You,
  Attia, Yang, Feng, Zhang, Ramrakhya, Jia, Nichols, Toshev, Yang, and
  Gan}]{yang2025ferretuilite}
Zhen Yang, Zi-Yi Dou, Di~Feng, Forrest Huang, Anh Nguyen, Keen You, Omar Attia,
  Yuhao Yang, Michael Feng, Haotian Zhang, Ram Ramrakhya, Chao Jia, Jeffrey
  Nichols, Alexander Toshev, Yinfei Yang, and Zhe Gan. 2025{\natexlab{b}}.
\newblock {Ferret-UI Lite}: Lessons from building small on-device {GUI} agents.
\newblock \emph{arXiv preprint arXiv:2509.26539}.

\bibitem[{Zhang et~al.(2025)Zhang, Chen, Lu, Zhou, Yu, and
  Zhang}]{zhang2025prune4web}
Jiayuan Zhang, Kaiquan Chen, Zhihao Lu, Enshen Zhou, Qian Yu, and Jing Zhang.
  2025.
\newblock {Prune4Web}: {DOM} tree pruning programming for web agent.
\newblock \emph{arXiv preprint arXiv:2511.21398}.

\bibitem[{Zheng et~al.(2024)Zheng, Gou, Kil, Sun, and Su}]{zheng2024seeact}
Boyuan Zheng, Boyu Gou, Jihyung Kil, Huan Sun, and Yu~Su. 2024.
\newblock {GPT-4V(ision)} is a generalist web agent, if grounded.
\newblock In \emph{Proceedings of the 41st International Conference on Machine
  Learning (ICML)}.

\bibitem[{Zhou et~al.(2025)Zhou, Zhang, Tong, Zhang, Chen, Kong, Cai, Liu,
  Wang, Zhou, and Hoi}]{zhou2026maiui}
Hanzhang Zhou, Xu~Zhang, Panrong Tong, Jianan Zhang, Liangyu Chen, Quyu Kong,
  Chenglin Cai, Chen Liu, Yue Wang, Jingren Zhou, and Steven Hoi. 2025.
\newblock {MAI-UI} technical report: Real-world centric foundation {GUI}
  agents.
\newblock \emph{arXiv preprint arXiv:2512.22047}.

\bibitem[{Zhou et~al.(2024)Zhou, Xu, Zhu, Zhou, Lo, Sridhar, Cheng, Ou, Bisk,
  Fried, Alon, and Neubig}]{zhou2024webarena}
Shuyan Zhou, Frank~F. Xu, Hao Zhu, Xuhui Zhou, Robert Lo, Abishek Sridhar,
  Xianyi Cheng, Tianyue Ou, Yonatan Bisk, Daniel Fried, Uri Alon, and Graham
  Neubig. 2024.
\newblock {WebArena}: A realistic web environment for building autonomous
  agents.
\newblock In \emph{International Conference on Learning Representations
  (ICLR)}.

\end{thebibliography}

\end{document}